\documentclass[11pt,a4paper]{article}
\usepackage[hyperref]{emnlp2020}
\usepackage{times}
\usepackage{latexsym}

\usepackage{color}
\usepackage{microtype}
\usepackage{graphicx}
\usepackage{booktabs}
\usepackage{multirow}
\usepackage{CJK}

\aclfinalcopy 
\def\aclpaperid{999} 

\title{Profile Consistency Identification for Open-domain Dialogue Agents}

\author{Haoyu Song$^1$, Yan Wang, Wei-Nan Zhang$^{1}$, Zhengyu Zhao$^1$, Ting Liu$^1$, Xiaojiang Liu \\
  $^1$Research Center for Social Computing and Information Retrieval\\
  Harbin Institute of Technology, Heilongjiang, China\\
  \texttt{\{hysong,wnzhang,zyzhao,tliu\}@ir.hit.edu.cn}\\
  \texttt{yanwang.branden@gmail.com}\ \ \texttt{xiaojiangliu84@hotmail.com} \\
  }

\date{}

\begin{document}
\maketitle
\begin{abstract}
Maintaining a consistent attribute profile is crucial for dialogue agents to naturally converse with humans.
Existing studies on improving attribute consistency mainly explored how to incorporate attribute information in the responses, but few efforts have been made to identify the consistency relations between response and attribute profile.
To facilitate the study of profile consistency identification, 
we create a large-scale human-annotated dataset with over 110K single-turn conversations and their key-value attribute profiles. Explicit relation between response and profile is manually labeled. We also propose a key-value structure information enriched BERT model to identify the profile consistency, and it gained improvements over strong baselines.
Further evaluations on downstream tasks demonstrate that the profile consistency identification model is conducive for improving dialogue consistency.
\end{abstract}

\section{Introduction}

Despite the recent advancements in assigning attribute profiles to dialogue agents~\citep{qian2018assigning,zhang2019dialogpt}, maintaining a consistent profile is still challenging for an open-domain dialogue agent. Existing works mainly emphasize the incorporation of attribute information in the generated responses~\citep{wolf2019transfertransfo,song-ijcai2019-721,zheng2019pre}. Although these models have improved the response consistency by explicitly modeling the profiles, they still face the consistency issue~\citep{welleck-etal-2019-dialogue}. One important reason is that they cannot identify the consistency relations between response and profile. 

\begin{figure}
\centering
\includegraphics[width=.99\columnwidth]{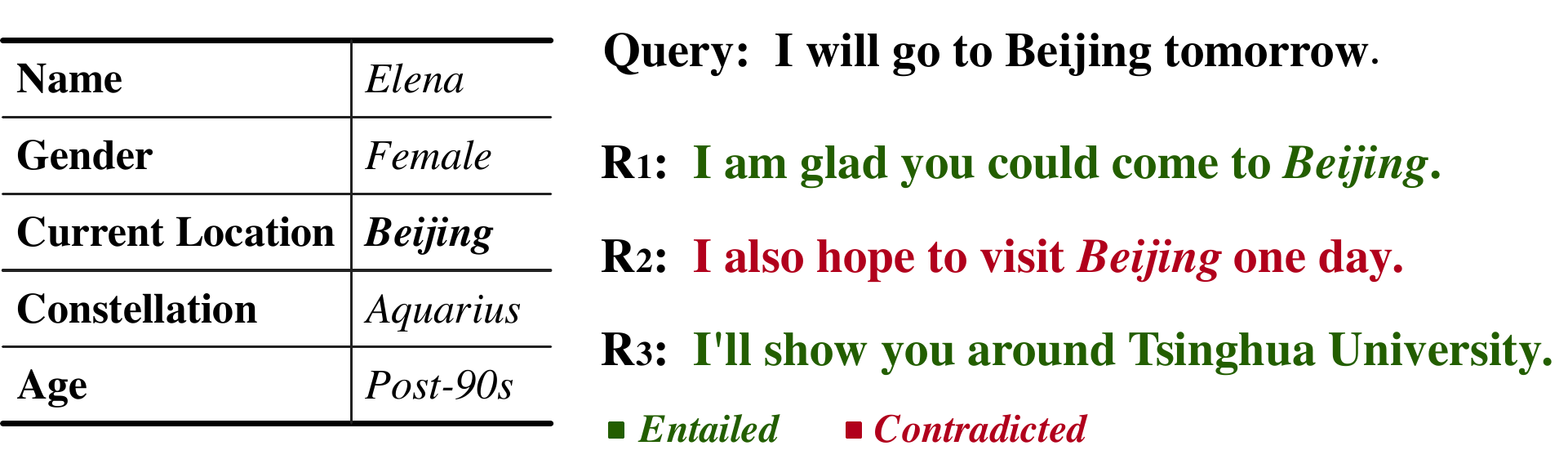}
\caption{ Left: the key-value attribute profiles of the dialogue agent. Right: a dialogue query with different responses that might be related to the attribute profiles.
Among these responses, $R_1$ entails the current location profile, while $R_2$ contradicts the profile. Although $R_3$ does not contain the attribute word {\it Beijing}, we could still understand $R_3$ entails the current location. 
}
\label{fig:1}
\end{figure}

As shown in Figure~\ref{fig:1}, the attribute word {\it Beijing} is incorporated in the first two responses, but only {$R_1$} is semantically consistent with the speaker's profile. For example, $R_2$ ``I also hope to visit {\it Beijing} one day.'' implies that the speaker has never been to Beijing, which contradicts the speaker's profile. On the other hand, although $R_3$ does not contain the attribute word {\it Beijing}, we could still infer from the words {\it Tsinghua University} that the speaker's current location entails the profile.
Existing studies~\citep{qian2018assigning,zheng2019personalized} train dialogue agents to produce plausible responses that contain attribute information, but still cannot teach agents to understand the differences of consistency relations in these responses.

\citet{welleck-etal-2019-dialogue} made an early step towards reducing the dialogue consistency identification to natural language inference (NLI)~\citep{bowman-etal-2015-large}, where they learn a mapping from two dialogue utterances to an entailment category. All utterances in~\citet{welleck-etal-2019-dialogue} are natural sentences from the PersonaChat dataset~\citep{zhang-etal-2018-personalizing}.
However, structured attribute profiles, such as key-value pairs, are ubiquitous in real-world dialogue systems~\citep{shum2018eliza}. Compared with natural sentences, structured profiles have fixed attribute keys from different domains and specific attribute values from limited candidates. The structure information is also essential to a better understanding of the profile.
To endow agents with the ability to identify structured profile consistency, we need a new dataset with fine-grained labels between response and profile, as well as a model that can leverage the structure information in the profile.


In this work, we introduce a human-annotated dataset, named Key-value Profile Identification (KvPI), with over 110K single-turn conversations and corresponding attribute profiles.
Three representative domains, gender, location, and constellation, are involved in the human annotation. We hire an annotation team to (1) label the relation ({\it entailed}, {\it contradicted}, or {\it irrelevant}) between each conversation and structured profile, and (2) find out the detailed attribute information in each response.

With the annotated KvPI dataset, we set up different baseline models, and propose a key-value structure information enriched BERT (KvBERT) model, which leverages dependency structures in profiles to enrich the contextual representations. Experimental results show that KvBERT obtains significant improvements over strong baselines.
We further test the KvBERT model on two downstream tasks, including a reranking task~\citep{welleck-etal-2019-dialogue} and a consistency prediction task~\citep{dziri-etal-2019-evaluating}.
Evaluation results show that (1) the KvBERT reranking improves response consistency, and (2) the KvBERT consistency prediction has a good agreement with human annotation.

Our contributions are summarized as below:
\begin{itemize}
  \item A KvPI dataset is introduced, which has over 110K fine-grained consistency annotations between responses and their key-value profiles.
  \item A KvBERT model is proposed for consistency identification, which gained significant improvements over strong baselines.
  \item Evaluations on downstream tasks show that the profile consistency identification model could be complementary to dialogue models.
\end{itemize}

\section{Dataset Preparation}
In this section, we describe the collection and annotation process of the KvPI dataset: (1) how we collect high-quality conversations and profiles; (2) how we define the consistency relations between responses and profiles; and (3) how we annotate consistency relations for the collected data.

\subsection{Data Collection}
To study the profile consistency identification problem, we use data from Weibo\footnote{https://en.wikipedia.org/wiki/Sina\_Weibo}, a popular and plentiful Chinese social media, in which people routinely respond to different posts and have publicly available profiles, such as gender and location.
We follow the protocol of the previous profile-based dialogue dataset~\citep{qian2018assigning,zheng2019personalized} to collect Weibo post-response pairs, together with users' available profiles. 
Here we filter out overly long or short pairs and finally obtain a tuple pool that contains about 30 million tuples, which are in a \{{\it profile}, {\it post}, {\it response}\} format. Each profile includes three popular attributes: {gender}, {location} and {constellation}, and organized in a key-value format. For instance,  \{{gender}: {\it female}, location:{\it Beijing}, constellation: {\it Aquarius}\}. This format is widely applied in real-world dialogue systems, such as~\citet{bowden2017combining},~\citet{shum2018eliza}, and ~\citet{pichl2018alquist}.

Since our goal is to identify explicit consistency relations between response and profile, we filter out the tuples whose response has no profile-related information by employing a pre-trained classifier and heuristic rules. Finally, we obtain about 150K profile-related tuples after filtering.

\subsection{Consistency Relations}
\label{sec:consistency_relations}
We define three types of consistency relation between the response and profile under the open-domain dialogue setting, which is different from the entailment categories in natural language inference~\citep{bowman-etal-2015-large,welleck-etal-2019-dialogue}:

\paragraph{Entailed} The response is exactly talking about the dialogue agent's attribute information, and the attribute is consistent with its key-value profile.

\paragraph{Contradicted} Although the response is talking about the dialogue agent's attribute information, it is contradicted to at least one of the given key-value pairs. For example, given the profile ``\{{location}: {\it Beijing}\}'', ``I am in Seattle'' is contradicted to the profile, while ``She lives in Seattle'' is not, because the latter is not talking about the dialogue agent's attribute.

\paragraph{Irrelevant} The response contains profile-related information, but the information does not reveal the dialogue agent's own attributes. As exemplified above, ``She lives in Seattle'' is irrelevant, rather than contradicted, to the dialogue agent's profile ``\{{location}: {\it Beijing}\}''. Another example is ``I'm interested in the history of Beijing''. Although there is the attribute word ``Beijing'', this response still does not reveal the dialogue agent's location.

\begin{table*}[]
\begin{CJK*}{UTF8}{gbsn}
\resizebox{\textwidth}{!}{%
\begin{tabular}{l|l|l|l|l|l}
\toprule
\textbf{Profile} &
  \textbf{Post} &
  \textbf{Response} &
  \textbf{Domain} &
  \textbf{Annotated Attribute} &
  \textbf{Label} \\ \midrule
\begin{tabular}[c]{@{}l@{}}Constell: Aries\\ Loc: Henan Anyang\\ Gender: Female\end{tabular} &
  \begin{tabular}[c]{@{}l@{}}Bro, are you also\\ a Scorpio?\\ \small{兄弟，你也天蝎啊?}\end{tabular} &
  \begin{tabular}[c]{@{}l@{}}I'm an Aries bullied by Scorpio\\ \\ \small{我是被天蝎欺负的白羊座}\end{tabular} &
  Constell &
  \{Constell: Aries\} &
  E \\ \midrule
\begin{tabular}[c]{@{}l@{}}Constell: Scorpio\\ Loc: Beijing\\ Gender: Female\end{tabular} &
  \begin{tabular}[c]{@{}l@{}}Too cold and you\\ girls will catch cold\\ \small{女孩子贪凉容易感冒}\end{tabular} &
  \begin{tabular}[c]{@{}l@{}}Are you confused? I'm not a girl!\\ I am a middle-aged woman!\\ \small{搞错了吧?人家不是女孩!是中年少女!}\end{tabular} &
  Gender &
  \{Gender: Female\} &
  E \\ \midrule
\begin{tabular}[c]{@{}l@{}}Constell: Leo\\ Loc:   Jiangsu\\ Gender: Male\end{tabular} &
  \begin{tabular}[c]{@{}l@{}}I am not here\\ \\ \small{我没在啊}\end{tabular} &
  \begin{tabular}[c]{@{}l@{}}Emm..I thought you came to Suzhou\\ \\ \small{嗯..还以为你来苏州了}\end{tabular} &
  Loc &
  \{Loc: Jiangsu Suzhou\} &
  E \\ \midrule
\begin{tabular}[c]{@{}l@{}}Constell: Virgo\\ Loc: Shaanxi Xi'an\\ Gender: Male\end{tabular} &
  \begin{tabular}[c]{@{}l@{}}Did you build it on\\ the site?\\ \small{你们工地建的?}\end{tabular} &
  \begin{tabular}[c]{@{}l@{}}Impossible! We are in Hancheng,\\ but the brand is in Xi'an\\ \small{不可能啦!我们在韩城，这块牌子在西安}\end{tabular} &
  Loc &
  \{Loc: Shaanxi Hancheng\} &
  C \\ \midrule
\begin{tabular}[c]{@{}l@{}}Constell: Taurus\\ Loc: Guangdong\\ Gender: Male\end{tabular} &
  \begin{tabular}[c]{@{}l@{}}I don't know how to\\ fix the computer\\ \small{我不知道怎么修电脑}\end{tabular} &
  \begin{tabular}[c]{@{}l@{}}Go to find your boyfriend ha ha\\ \\ \small{找你男人去哈哈}\end{tabular} &
  Gender &
  None &
  I \\ \midrule
\begin{tabular}[c]{@{}l@{}}Constell: Gemini\\ Loc: Fujian\\ Gender: Female\end{tabular} &
  \begin{tabular}[c]{@{}l@{}}What kind of food\\ do you want?\\ \small{你想要什么好吃的呀?}\end{tabular} &
  \begin{tabular}[c]{@{}l@{}}I want the Taiwan soy-braised pork\\ \\ \small{想吃台湾红烧肉}\end{tabular} &
  Loc &
  None &
  I \\ 
\bottomrule
\end{tabular}%
}
\end{CJK*}
\caption{Examples of KvPI dataset. These sentences are in Chinese, and we translated them into English. Constell and Loc are short for constellation and location. E, C, I denote Entailed, Contradicted, and Irrelevant, respectively.}
\label{tab:1}
\end{table*}

\subsection{Human Annotation}
The definitions in Sec~\ref{sec:consistency_relations} are also applied in the human annotation process.
We hire an annotation team to (1) review whether the response is profile-related, and (2) annotate the fine-grained information, including consistency labels, domains, and detailed attributes in each response. To ensure quality, each tuple is annotated by three people, and the annotation process lasts nearly four months.

In the annotation process, about 10K tuples are filtered out due to no profile-related information in their responses, and we obtain 140K valid tuples with explicit annotations of consistency relation.

\subsection{Quality Control}
To control the quality of the annotated dataset, we introduce different verification methods:

First, in the annotation process, we review 200 randomly sampled tuples every 10,000 annotations. We assign a ``gold'' label to each tuple and then decided whether the whole annotation batch should be accepted or re-annotated according to the disagreement rate. With tolerance to the different understandings of the dialogue response, we set an empirical acceptance threshold of disagreement rate to 10\%. For the majority of annotated batches, the disagreement rate varies from 3\% to 7\%. 

The second verification is conducted by paid annotators. Each consistency label is verified by two annotators. The tuples with a low inter-annotator agreement in their labels are directly discarded from the final dataset. Finally, we obtain 118,540 tuples in the KvPI dataset.

From the final dataset, we randomly sampled 2,000 profile-response pairs to two new annotators. These pairs are also annotated as entailed, contradicted, and irrelevant, as in the completed annotation process. Following~\citet{bowman-etal-2015-large}, we calculated the Fleiss' Kappa among the previous labels and two new labels and obtained a kappa of 0.857, which means {\it almost perfect agreement}~\citep{landis1977measurement}. This result shows that the completed annotation is of good quality.

\section{The KvPI Dataset}
We present some examples of the final KvPI dataset in Table~\ref{tab:1}.
The dataset, together with trained models, will be open-sourced for public usage.

\begin{table*}[]
\resizebox{\textwidth}{!}{%
\begin{tabular}{@{}l|ccc|ccc|cccc@{}}
\toprule
  \textbf{Domains}&
  \textbf{Entail} &
  \textbf{Contr} &
  \textbf{Irrelv} &
  \textbf{Len(E)} &
  \textbf{Len(C)} &
  \textbf{Len(I)} &
  \textbf{Train} &
  \textbf{Valid} &
  \textbf{Test} &
  \textbf{Overall} \\ \midrule
Gender   & 8,270  & 6,858  & 16,201 & 20.5 & 20.6 & 20.8 & 25,329 & 3,000  & 3,000  & 31,329  \\
Location & 18,468 & 17,777 & 28,759 & 15.8 & 15.9 & 17.5 & 53,004 & 6,000  & 6,000  & 65,004  \\
Constell & 6,376  & 6,365  & 9,466  & 14.5 & 14.6 & 16.7 & 18,207 & 2,000  & 2,000  & 22,207  \\ \midrule
Total    & 33,114 & 31,000 & 54,426 & 16.7 & 16.7 & 18.3 & 96,540 & 11,000 & 11,000 & 118,540 \\ \bottomrule
\end{tabular}%
}
\caption{Basic statistics of the KvPI dataset. We depict the statistics from the perspective of three domains. Entail, Contr, and Irrelv are short for Entailment(E), Contradiction(C), and Irrelevant(I), respectively.}
\label{tab:2}
\end{table*}

\subsection{Dataset Organization}
\label{sec:2.2}
The KvPI dataset consists of single-turn conversations and profiles, labeled as entailed, contradicted, or irrelevant. Attributes in the dataset profiles come from three domains, including {gender}, {location}, and {constellation}. The profile is organized in a key-value format, for example, \{{gender}: {\it female}, {location}: {\it Beijing}, {constellation}: {\it Leo}\}.

\paragraph{Gender} This domain includes responses that have evidence indicating they are from men or women. Both explicit gender evidence, such as ``I am a girl'', and implicit gender evidence, such as ``I'm hanging out with my boyfriend'', are included.

\paragraph{Location} This domain includes responses talking about the locations. Besides the accurate matching of location, data in this domain also needs common sense reasoning, such as whether a city belongs to a province, as shown in the third example in Table~\ref{tab:1}.

\paragraph{Constellation} This domain includes different responses that talk about the constellation. A good number of the responses contain more than one constellation word.

Both entailed and irrelevant cases in the KvPI dataset are directly obtained from the annotation results. To balance the number of cases in each relation, we collect the contradicted cases from two sources: (1) the annotated contradicted tuples, and (2) the rewritten entailed tuples.
Possible reasons for the originally contradicted cases are that users may forget to update their profiles, or they are intended to present different information about themselves.
Data from the first source accounts for about two-thirds of the total contradicted cases.
The other part comes from entailed cases. Their profiles have been rewritten to different attributes, with a minimal edit-distance principle, so that they turn into contradicted. Cases from this source are treated as new data in the annotation process. Unqualified rewritten data is discarded.

\subsection{Statistics}
Table~\ref{tab:2} summarizes the main statistics of the KvPI dataset. The first and third groups in Table~\ref{tab:2} count the number of unique tuples in the dataset. 
Here a tuple refers to a group of data consisting of a key-value profile, a post, a dialogue response, as well as the corresponding domain, the annotated attribute, and the label of consistency relation. 
The tuple examples can be seen in Table~\ref{tab:1}. 
For the second group, it only calculates the average number of tokens in the dialogue responses.

\section{Profile Consistency Identification}

\begin{figure*}
  \centering
  \includegraphics[width=.98\textwidth]{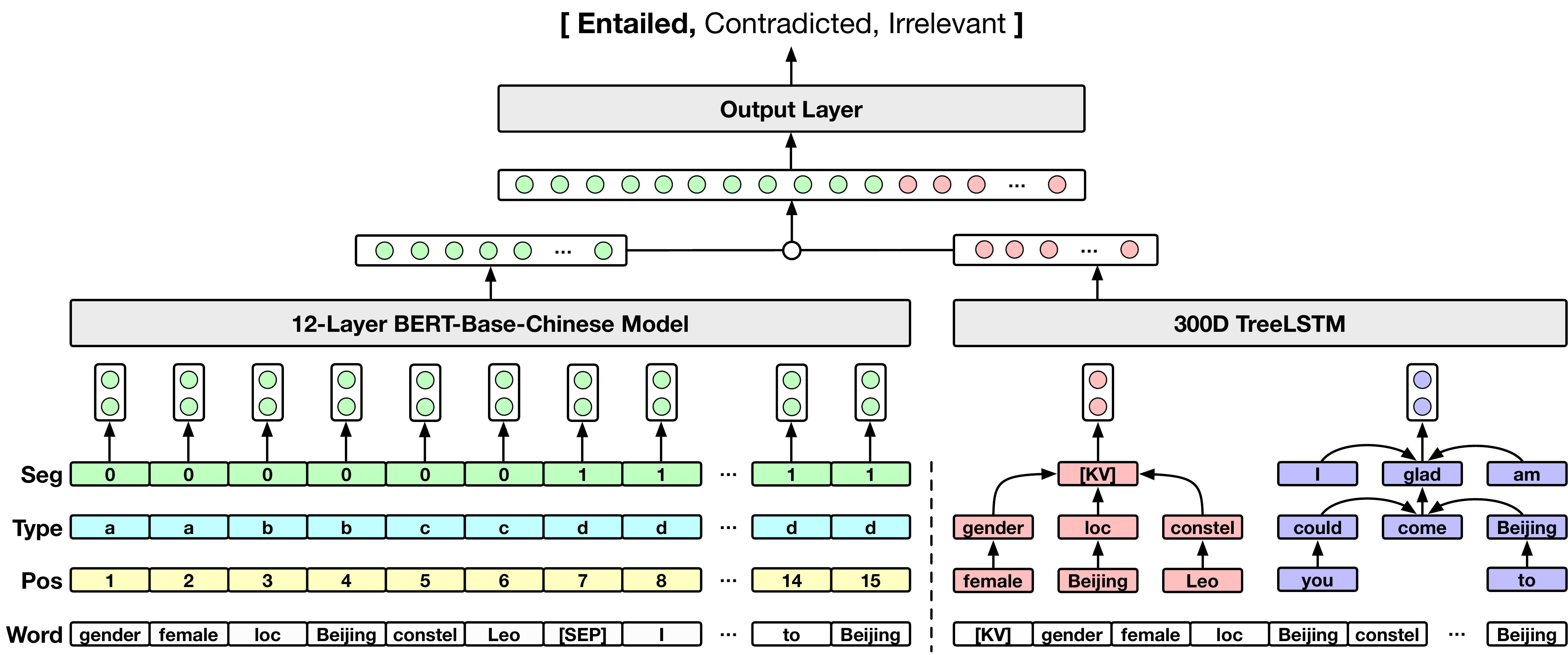}
  \caption{The overall framework of the KvBERT model. Examples in this figure: the key-value profile is \{ gender: {\it female}, location: {\it Beijing}, constellation: {\it Leo}\}, and the dialogue response is ``{\it I am glad you could come to Beijing}''.
  }
  \label{fig:2}
\end{figure*}

\subsection{Problem Definition}
\label{sec:definition}
To equip dialogue agents with the ability to identify consistency, we need to build a profile consistency identification model. This model learns to identify the relation of \{entailed, contradicted, irrelevant\} between a (profile, response) pair. 
Formally, our goal is to learn a mapping function $\mathcal F$, and $\mathcal F(P,R)\in\{e,c,i\}$, where $P$=$\{k_1:v_1,...,k_n:v_n\}$, $R=w_1,w_2,...,w_m$. Here $P$ denotes the key-value profile, and $R$ denotes the response with $m$ words. $e$, $c$, $i$ denote the consistency relations.

\subsection{Motivation}
\label{sec:motivation}

The main challenge of identifying profile consistency lies in how to model the key-value profiles effectively. Such structured profiles have a common dependency structure, which differs from the natural sentences.
For example, from the profile \{ gender: {\it female}, location: {\it Beijing}, constellation: {\it Leo}\}, we can clearly see three dependency relations: female $\rightarrow$ gender, Beijing $\rightarrow$ location, and Leo $\rightarrow$ constellation.  Moreover, gender, location, and constellation will define the information in the {\it kv-profile}. Here we can see a hierarchical structure of the key-value profiles, as illustrated in Figure~\ref{fig:2}. More importantly, no matter how the values change, this structure will stay unchanged. 

Although large pre-trained models such as BERT implicitly capture dependency information more or less~\citep{clark-etal-2019-bert}, we argue that such implicit syntactic information may not be enough to support a powerful contextual representation for reasoning on the highly structured key-value profiles, according to the meaningless dependency parsing results generated by BERT on the structured profiles.

These observations motivate us to incorporate the explicit structure of profiles directly. To this end, we design the KvBERT, which integrates both language representation from BERT and structure representation from tree-LSTM~\citep{pmlr-v37-zhub15}.

\subsection{Model Brief}
\label{sec:brief}
Figure~\ref{fig:2} shows the overall framework of the KvBERT model. 
On the BERT side, we linearize the key-value pairs into a sequence and treating the responses as another sequence\footnote{Our data collection scheme ensures that all responses contain profile information, which frees the modeling of post.}. The input embedding is the sum of four embeddings, including an additional type embedding~\citep{2019TabFactA} to inform the model of different key-value pairs, as shown in Figure~\ref{fig:2}. Here we omit the well-known formulations of BERT~\citep{devlin-etal-2019-bert} for brevity. We can get a contextual representation for the linearized sequence through the BERT model.

On the tree-LSTM side, the profiles are parsed to predefined structure, as discussed in Sec~\ref{sec:motivation}. An example of this structure can be seen in the red part of the Figure~\ref{fig:2}. In parallel, the responses are passed to a trained parser to fetch the dependency structure. Then the tree-LSTM encodes two structures to corresponding embeddings. Three operations are performed to aggregate information from two embeddings: element-wise multiplication, element-wise difference, and concatenation. The aggregated embedding is followed up by a linear layer to form the final structure representation.

At last, the sentence representation and structure representation are concatenated to form the joint representation for the final linear output layer.

\subsection{The Dependency Structures}
In our model, the dependency structure for profiles is predefined, and for the response, it is obtained from a trained parser.
To complete the structure in the profile, we add a special [KV] token on the top of the dependency structure of the profile. As a result, the [KV] token aggregates information from its child key-value nodes.
In contrast, there is no universal dependency structure in the responses. To obtain the structures in the responses, we trained a parser on CDT5.0 (Chineses dependency treebank), achieving 90.72\% and 88.38\% unlabeled and labeled attachment score. All structure predictions are made in the data preprocessing stage.

A tree-LSTM unit encodes multiple child units or multiple descendant units in a recursive process. Due to the length limit, we recommend readers to get the details from~\citet{pmlr-v37-zhub15}. For both the predefined structures and the parsed structures, we apply the same depth-first encoding strategy, from every leaf node to the root node, to aggregate the structure information.

\section{Experiments}
In this section, we first evaluate the performance of the proposed KvBERT model on identifying profile consistency. After that, we test the trained KvBERT model on two downstream tasks, including a reranking task and a consistency prediction task, to analyze how well the proposed approach performs under practical applications.

\begin{table*}[]
\resizebox{\textwidth}{!}{%
\begin{tabular}{@{}lllllllll@{}}
\toprule
\multicolumn{1}{l|}{\textbf{Metrics\ \ \ \ \ \ Domains}} &
  \multicolumn{4}{c|}{\textbf{KvPI Test Set}} &
  \multicolumn{4}{c}{\textbf{Gender}} \\ \cmidrule(l){2-9} 
\multicolumn{1}{l|}{\textbf{Models}}    & acc        & entail-f1 & contr-f1  & \multicolumn{1}{l|}{irrelv-f1} & acc        & entail-f1 & contr-f1  & irrelv-f1 \\ \midrule
\multicolumn{1}{l|}{SVM+uni+bi}         & 61.3 (14)  & 73.6 (18)  & 55.9 (5.5) & \multicolumn{1}{l|}{41.5 (3.5)} & 53.0 (17)  & 69.0 (17)  & 42.2 (78)  & 17.8 (98)  \\
\multicolumn{1}{l|}{SVM+uni+bi+overlap} & 68.7 (8.5) & 76.2 (13)  & 65.1 (32)  & \multicolumn{1}{l|}{50.3 (13)}  & 60.0 (44)  & 73.9 (39)  & 48.9 (47)  & 16.7 (89)  \\
\multicolumn{1}{l|}{ESIM-template}      & 83.1 (4.8) & 81.7 (7.1) & 85.8 (0.8) & \multicolumn{1}{l|}{79.6 (1.1)} & 76.8 (5.9) & 70.6 (6.5) & 85.1 (3.3) & 62.8 (19)  \\
\multicolumn{1}{l|}{ESIM-kv}            & 83.7 (0.8) & 82.0 (4.1) & 86.3 (1.7) & \multicolumn{1}{l|}{80.6 (1.7)} & 77.9 (1.3) & 72.7 (1.6) & 85.7 (2.4) & 63.7 (11)  \\
\multicolumn{1}{l|}{GPT-template}       & 86.5 (0.2) & 88.1 (1.2) & 86.3 (1.9) & \multicolumn{1}{l|}{83.9 (2.5)} & 80.0 (2.5) & 87.0 (4.1) & 75.1 (6.9) & 68.0 (15)  \\
\multicolumn{1}{l|}{GPT-kv}             & 86.4 (0.5) & 88.2 (0.9) & 86.1 (2.2) & \multicolumn{1}{l|}{83.8 (2.2)} & 80.1 (2.6) & 87.2 (1.3) & 74.9 (6.9) & 68.3 (11)  \\
\multicolumn{1}{l|}{BERT-template}      & 87.1 (0.4) & 88.7 (1.4) & 86.7 (1.7) & \multicolumn{1}{l|}{84.9 (1.3)} & 81.4 (1.5) & 87.9 (1.0) & 77.2 (0.5) & 70.5 (6.0) \\
\multicolumn{1}{l|}{BERT-kv}            & 88.0 (1.7) & 89.5 (2.2) & 87.2 (0.8) & \multicolumn{1}{l|}{86.2 (2.2)} & 80.3 (7.6) & 87.5 (2.9) & 75.6 (12)  & 68.9 (13)  \\
\multicolumn{1}{l|}{TableBERT}          & 88.6 (2.1) & 89.8 (3.3) & 88.1 (4.5) & \multicolumn{1}{l|}{87.1 (1.7)} & 81.7 (0.9) & 87.4 (3.8) & 77.9 (8.2) & 74.0 (8.6) \\ \midrule
\multicolumn{1}{l|}{\bf KvBERT (Ours)} &
  \textbf{91.7 (1.3)} &
  \textbf{93.3 (1.7)} &
  \textbf{91.0 (1.4)} &
  \multicolumn{1}{l|}{\textbf{90.1 (0.8)}} &
  \textbf{85.9 (2.1)} &
  \textbf{91.3 (1.2)} &
  \textbf{81.4 (3.6)} &
  \textbf{77.8 (2.9)} \\ \midrule
                                        &            &            &            &                                 &            &            &            &            \\ \midrule
\multicolumn{1}{l|}{\textbf{Metrics\ \ \ \ \ \ Domains}} &
  \multicolumn{4}{c|}{\textbf{Location}} &
  \multicolumn{4}{c}{\textbf{Constellation}} \\ \cmidrule(l){2-9} 
\multicolumn{1}{l|}{\textbf{Models}}    & acc        & entail-f1 & contr-f1  & \multicolumn{1}{l|}{irrelv-f1} & acc        & entail-f1 & contr-f1  & irrelv-f1 \\ \midrule
\multicolumn{1}{l|}{SVM+uni+bi}         & 62.4 (47)  & 66.1 (72)  & 59.7 (41)  & \multicolumn{1}{l|}{59.7 (22)}  & 49.4 (2.9) & 66.1 (5.4) & 22.6 (77)  & 7.5 (98)   \\
\multicolumn{1}{l|}{SVM+uni+bi+overlap} & 69.2 (30)  & 58.7 (94)  & 76.4 (15)  & \multicolumn{1}{l|}{71.4 (23)}  & 74.1 (36)  & 78.1 (25)  & 41.3 (99)  & 87.1 (5.7) \\
\multicolumn{1}{l|}{ESIM-template}      & 85.2 (0.9) & 87.7 (2.2) & 85.4 (0.5) & \multicolumn{1}{l|}{82.3 (2.4)} & 88.5 (0.0) & 82.6 (3.3) & 88.5 (1.4) & 94.2 (1.4) \\
\multicolumn{1}{l|}{ESIM-kv}            & 85.5 (0.8) & 87.9 (0.8) & 85.5 (2.5) & \multicolumn{1}{l|}{82.8 (0.0)} & 87.6 (7.8) & 83.0 (9.0) & 88.6 (7.9) & 92.0 (9.1) \\
\multicolumn{1}{l|}{GPT-template}       & 87.7 (1.6) & 87.5 (1.7) & 90.1 (7.0) & \multicolumn{1}{l|}{84.9 (1.6)} & 92.2 (1.4) & 91.5 (2.5) & 88.2 (2.1) & 96.9 (2.4) \\
\multicolumn{1}{l|}{GPT-kv}             & 87.7 (1.3) & 87.6 (1.4) & 90.3 (4.6) & \multicolumn{1}{l|}{84.8 (5.7)} & 91.5 (1.4) & 90.9 (1.7) & 87.3 (1.4) & 96.6 (2.9) \\
\multicolumn{1}{l|}{BERT-template}      & 89.9 (2.0) & 89.9 (1.5) & 91.2 (1.0) & \multicolumn{1}{l|}{89.2 (2.0)} & 92.5 (0.5) & 91.9 (1.6) & 88.4 (1.4) & 97.2 (1.3) \\
\multicolumn{1}{l|}{BERT-kv}            & 89.9 (1.4) & 88.6 (2.9) & 91.2 (0.9) & \multicolumn{1}{l|}{89.8 (1.4)} & 92.1 (4.5) & 91.7 (3.7) & 87.9 (7.4) & 97.0 (3.3) \\
\multicolumn{1}{l|}{TableBERT}          & 90.2 (1.9) & 90.1 (2.8) & 91.4 (2.9) & \multicolumn{1}{l|}{89.5 (0.5)} & 92.9 (1.4) & 92.5 (4.3) & 89.9 (1.7) & 97.2 (0.5) \\ \midrule
\multicolumn{1}{l|}{\bf KvBERT (Ours)} &
  \textbf{92.8 (1.7)} &
  \textbf{93.1 (1.2)} &
  \textbf{93.4 (2.5)} &
  \multicolumn{1}{l|}{\textbf{91.7 (2.6)}} &
  \textbf{94.5 (1.2)} &
  \textbf{94.2 (2.5)} &
  \textbf{91.5 (2.8)} &
  \textbf{97.8 (1.9)} \\ \bottomrule
\end{tabular}%
}
\caption{Evaluation results on the KvPI dataset. In brackets is the standard deviation of three runs, scaled by $10^{-3}$.}
\label{tab:3}
\end{table*}

\subsection{Experiment Settings}
In our experiments, we train the KvBERT based on the 12-layer BERT-Base-Chinese model, with an embedding and hidden dimension of 768. For the tree-LSTM, we set embedding size to 300 and output dimension to 50. The dimension of the final representation is 818. The tree-LSTM is firstly pre-trained on the KvPI dataset for 13 epochs and then jointly finetuned with BERT representations for 3 epochs. The KvBERT model is implemented in {\it PyTorch}. More setting details are in the appendix.

\subsection{Identifying Profile Consistency}
We compare the performance of a variety of baseline models on identifying profile consistency:

\paragraph{Feature-based classifier} Our goal of setting this baseline was to better understand the difficulty of identifying profile consistency, rather than necessarily a state-of-the-art model. Here we choose SVM as the classifier, with unigram features and bigram features, i.e., {\bf SVM+uni+bi}. Additionally, the overlaps between profile values and responses are extracted as another feature, which is the {\bf SVM+uni+bi+overlap}.

\paragraph{Rnn-based NLI model} ESIM~\citep{chen-etal-2017-enhanced} is a powerful natural language inference model, which enhanced the interactions in the LSTM. This model was applied in~\citet{welleck-etal-2019-dialogue} and achieved the best results. Therefore, we set {\bf ESIM} as the rnn baseline for our experiments.

\paragraph{Pretrained models} Large pre-trained transformers have been shown effective for natural language understanding tasks. We choose the Generative Pre-trained Transformer, i.e. {\bf GPT}~\citep{radford2018improving}, and Bidirectional Encoder Representations from Transformers, i.e. {\bf BERT}~\citep{devlin-etal-2019-bert} as our pre-trained baselines.~\citet{2019TabFactA} proposed a {\bf TableBERT} model, which models structured table information within the BERT framework. We take this model as another pre-trained baseline.
We did not explore other pre-trained models in this work, due to the expensive computational costs in preparing their Chinese models. We leave the exploration as future work.

Considering the previous works are designed for natural sentences, for the sake of a fair and thorough comparison, we use templates to convert the key-value profiles into natural sentences. The methods experimented on the converted dataset is marked by a suffix ``{\bf -template}''. And the comparative experiments on the original KvPI dataset are marked by ``{\bf -kv}'', which linearizes the original key-value profiles, the same as Sec~\ref{sec:brief}. Other models are directly evaluated on the original KvPI dataset.

For evaluations, despite the whole dataset that includes all three domains, we are also interested in the model's performance on each individual domain\footnote{Models on each domain are trained separately.}. We use {\bf accuracy} (acc), which has been widely applied in the natural language inference tasks, to measure the overall performance on each domain. To have a better look at the model's ability on identifying different consistency relations, we also calculate the f1-score of three relations under the same domain, i.e., {\bf entail-f1}, {\bf contr-f1}, and {\bf irrelv-f1}. The accuracy and f1-score are calculated by using toolkits from {\it sklearn}.

We report the averaged best results of three different runs on each domain in Table~\ref{tab:3}. With the explicit modeling of profile structures, our KvBERT achieves the best performance on all metrics across all domains. More importantly, KvBERT is the only model whose all metrics are over 90\% on the KvPI test set, especially compared with strong pre-trained baselines. Moreover, we also obtain 3.1\% absolute improvements on the overall accuracy to the latest TableBERT model~\citep{2019TabFactA}.

We noticed an interesting phenomenon between the BERT-kv and BERT-template: the performance of BERT-template on all three individual domains are better than the BERT-kv's. Nevertheless, on the overall test set, their performances are entirely reversed. One possible reason is that the converted profile loses the structure information. Even for the powerful BERT model, this kind of information still affects the overall performance.

\subsection{Testing on Downstream Tasks}

Now that the KvBERT achieves good performance on the KvPI dataset, we want to test the abilities of the proposed approach further. Similar to the evaluations of pre-trained language models, we evaluate the abilities of our trained KvBERT model on two downstream tasks, with the assistance of human annotation.

Here we consider two types of dialogue models, i.e., retrieval model and generation model. We test the KvBERT on two tasks: (1) Reranking the top 20 responses from a retrieval model, to see whether the profile consistency is improved~\citep{welleck-etal-2019-dialogue}. (2) Given the responses from state-of-the-art generative dialogue models, to see how well the KvBERT's consistency prediction agrees with the human annotation~\citep{dziri-etal-2019-evaluating}.

To build the testbeds of different dialogue models, we use the Chinese PersonalDialog~\citep{zheng2019personalized} dataset, which consists of over 20 million dialogues from Weibo, together with diversified profile traits and interests tags of the user.

Further, we manually create 100 test samples for each domain, and we abbreviate the test set in this section as {\bf Gen} (gender), {\bf Loc} (location), and {\bf Con} (constellation). Thus there are 300 test samples in total. Each test sample consists of a (profile, post) pair, where the attribute keys are the same as in the KvPI dataset. Moreover, we confirm that these posts will lead to domain-specific responses.

\begin{table}[]
\resizebox{\columnwidth}{!}{%
\begin{tabular}{@{}llccc@{}}
\toprule
\multicolumn{2}{l|}{\textbf{Domains}}              & Entail (\%) & Contr (\%)  & Irrelv (\%) \\ \midrule
\multicolumn{1}{l|}{\multirow{2}{*}{\textbf{Gen}}} & \multicolumn{1}{l|}{top-1} & 56.0 / 57.0 & 9.0 / 9.0   & 35.0 / 34.0 \\
\multicolumn{1}{l|}{} & \multicolumn{1}{l|}{top-5} & 43.2 / 51.0 & 9.2 / 7.8   & 47.6 / 41.2 \\ \midrule
\multicolumn{1}{l|}{\multirow{2}{*}{\textbf{Con}}} & \multicolumn{1}{l|}{top-1} & 22.0 / 30.0 & 20.0 / 6.0  & 58.0 / 64.0 \\
\multicolumn{1}{l|}{} & \multicolumn{1}{l|}{top-5} & 29.8 / 32.4 & 18.4 / 8.2  & 51.8 / 59.4 \\ \midrule
\multicolumn{1}{l|}{\multirow{2}{*}{\textbf{Loc}}} & \multicolumn{1}{l|}{top-1} & 10.0 / 11.0 & 33.0 / 11.0 & 57.0 / 78.0 \\
\multicolumn{1}{l|}{} & \multicolumn{1}{l|}{top-5} & 8.6 / 12.2  & 34.0 / 11.6 & 57.4 / 76.2 \\ \bottomrule
\end{tabular}%
}
\caption{Human annotations for the profile consistency of the retrieved responses {\bf before / after} reranking. }
\label{tab:4}
\end{table}

\subsubsection*{Task I: Reranking Retrieved Responses}

We build the retrieval model using {\it pylucene}. To retrieve responses, we index both profiles and responses in the PersonalDialog dataset, with weights 0.15 and 0.85 for the profile and response, respectively. We retrieve the top 20 candidate responses for each testing sample, and then these responses are reranked by the trained KvBERT model, according to the order {\it Entailed} \textgreater {\it Irrelevant} \textgreater {\it Contradicted}. Within the same category, the model confidence will determine the order.
Among the 20 responses from one test sample, the top 5 responses, both before and after reranking, are annotated by three people into entailed (Entail), contradicted (Contr), and irrelevant (Irrelv).

\begin{table*}[]
\resizebox{\textwidth}{!}{%
\begin{tabular}{@{}l|ccc|c|ccc|c|cccc@{}}
\toprule
\multirow{2}{*}{} & \multicolumn{4}{c|}
{\textbf{Gender}} & \multicolumn{4}{c|}{\textbf{Constellation}} & \multicolumn{4}{c}{\textbf{Location}}                     \\ \cmidrule(l){2-13} 
                  & ent-f1 & con-f1 & irr-f1 & $\kappa$ & ent-f1   & con-f1   & irr-f1   & $\kappa$  & ent-f1 & con-f1 & \multicolumn{1}{c|}{irr-f1} & $\kappa$ \\ \midrule
AR  & 97.0\% & 79.2\% & 69.6\% & 0.777 & 94.7\% & 78.8\% & 72.4\% & 0.744 & 94.3\% & 96.3\% & \multicolumn{1}{c|}{90.9\%} & 0.913 \\
TT  & 96.7\% & 75.0\% & 66.7\% & 0.736 & 91.4\% & 72.2\% & 65.5\% & 0.659 & 90.7\% & 96.1\% & \multicolumn{1}{c|}{69.6\%} & 0.847 \\ \bottomrule
\end{tabular}%
}
\caption{F1-score of model prediction against human annotation, with Cohen's Kappa to measure the agreements.
}
\label{tab:5}
\end{table*}

We report the statistics of annotation results in Table~\ref{tab:4} and show some reranking examples in the appendix. Besides the entailed responses, the irrelevant ones are more acceptable than the contradicted ones.
As we can see, the KvBERT reranking improves profile consistency, 
either by increasing the rate of entailment or by decreasing the rate of contradiction. The annotation results also concur with our intuition: selecting a proper response with the right location is difficult for the retrieval models.

\subsubsection*{Task II: Consistency Prediction}
In this task, we want to test how well the KvBERT's consistency prediction agrees with the human annotation on generated responses.
We implement two state-of-the-art profile-based dialogue generation models as the testbeds for this task, including the TransferTransfo~\citep{wolf2019transfertransfo} ({\bf TT}) and {AttentionRouting}~\citep{zheng2019pre} ({\bf AR}). Both models are based on pre-trained transformers. First, we pre-train two models on 4G Chinese news data and finetune them on the PersonalDialog dataset.
Then we use the trained models to generate responses on the test data {Gen}, {Con}, {Loc}, respectively. 

The collected responses are annotated into entailed, contradicted, and irrelevant by three annotators. The annotation instructions are the same as in Sec~\ref{sec:consistency_relations}. In parallel, the KvBERT also predicts the relations between each profile and response.

We first report the f1-score of model prediction against the human annotation in Table~\ref{tab:5}. We also report Cohen's Kappa~\citep{cohen1960coefficient} between human annotations and model prediction to measure their agreements directly. All metrics are calculated by {\it sklearn}.
From the f1-scores, we can see that the model predictions are similar to the human annotations in most cases. And the $\kappa$ coefficients show the good agreements more directly, where $\kappa$ between 0.6 and 0.8 indicates {\it substantial agreement}, and over 0.8 indicates {\it almost perfect agreement}~\citep{landis1977measurement}.

Responses from the generative models are in a different distribution from the training data, due to the model learning process. Still, the KvBERT obtains good agreements with humans. It shows the good generalization ability of the proposed method.

\subsection{Effects of the Structure Information}
\begin{figure}
  \centering
  \includegraphics[width=0.9\columnwidth]{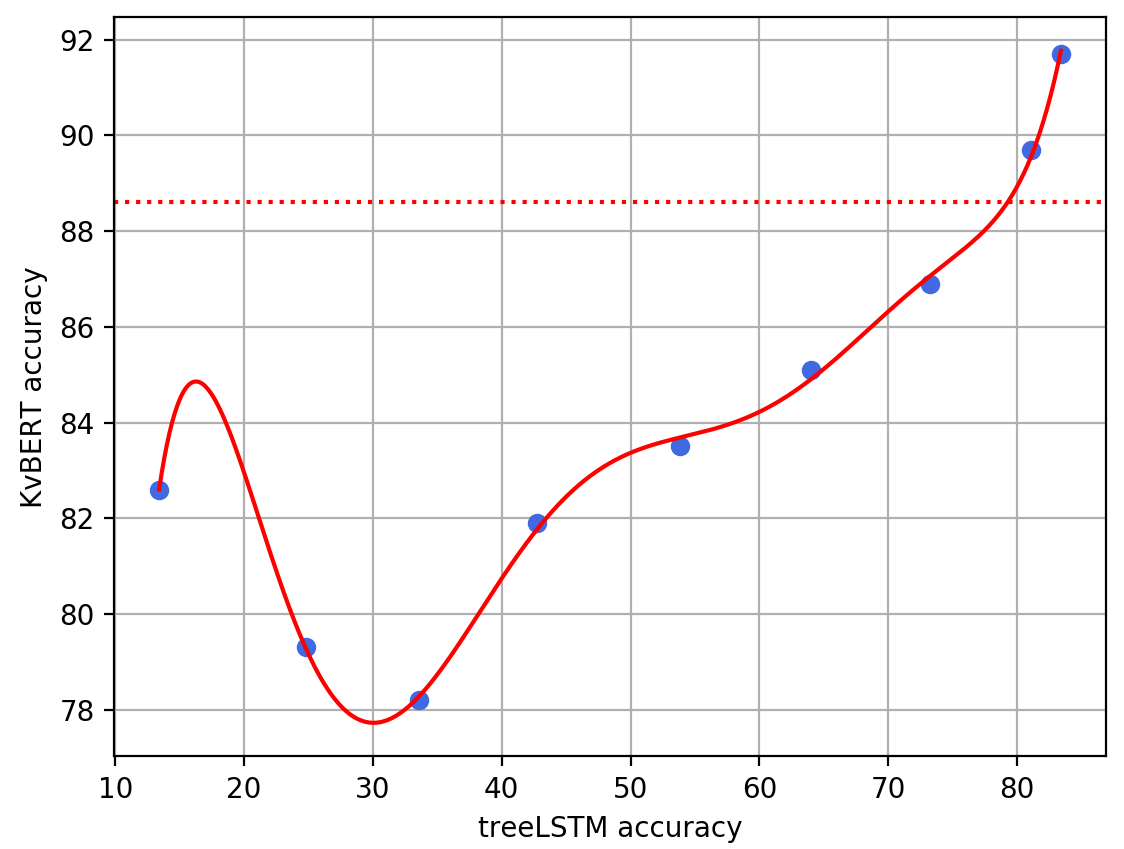}
  \caption{The red dashed line in the horizontal direction is the TableBERT accuracy, which has no structural information. The depicted curve is fitted by a seventh-degree polynomial.
  }
  \label{fig:structure}
\end{figure}

Another important question is whether the structure information is always helpful. To analyze this,
we sampled 9 treeLSTM checkpoints, with accuracy on the KvPI test set from 13.4\% to 83.4\%. The accuracy could be an indicator of how well the structure information has been captured. Then we trained 9 different KvBERT models with initialization from the 9 treeLSTMs and get final accuracies on the KvPI test set. We depict the treeLSTM accuracy and KvBERT accuracy, as well as a seventh-degree polynomial curve fitting the 9 data points, in Figure~\ref{fig:structure}. And there is a performance baseline shown by the dashed horizontal line, which has no structure information.

As we can see, not all the structural information contributes to the final performance. When the treeLSTM is at a low accuracy, the performance of the KvBERT model is inferior to that of the baseline model. Especially when the accuracy of treeLSTM is lower than 30\%, the final performance is even getting worse when the accuracy of treeLSTM grows. And only when the accuracy of treeLSTM is higher than about 80\%, can the final performance be improved, as illustrated in Figure~\ref{fig:structure}.

\subsection{Reproducibility} The code, data, and trained model are available at https://github.com/songhaoyu/KvPI.

\section{Related Work}
This work is closely related to the researches in natural language inference~\citep{bowman-etal-2015-large}. NLI aims to determine whether a natural language hypothesis can be inferred from a natural language premise~\citep{bowman-etal-2015-large,williams-etal-2018-broad,khot2018scitail,welleck-etal-2019-dialogue}. Besides the natural language evidence,~\citet{suhr-etal-2017-corpus} and~\citet{suhr-etal-2019-corpus} proposed to use images as the evidence for statement verification under the multi-modal setting.
A more recent related work is the~\citet{2019TabFactA}, who proposed to use semi-structured Wikipedia tables as evidence. The difference between our work and~\citet{2019TabFactA} is noticeable: open-domain dialogues have unique language patterns, and the key-value profiles are highly structured, as analyzed in Sec~\ref{sec:motivation}. To the best of our knowledge, this is the first work that explores the identification of consistency between dialogue responses and structured profiles.

Another line of research related to this work is the personalized dialogue generation task~\citep{zhang-etal-2018-personalizing,qian2018assigning,zheng2019personalized,song2020nli,song-etal-2020-generate}. This task seeks to improve personality consistency by incorporating persona information in the generated responses. For this purpose, several personalized dialogue datasets have been introduced in recent years, such as PersonaChat~\citep{zhang-etal-2018-personalizing} and PersonalDialog~\citep{zheng2019personalized}. These datasets successfully inform models of how to incorporate attribute related information in the responses, but still can not teach models how to identify the consistency relations between their response and profile.

\section{Conclusion and Discussion}

In this work, we introduce a large-scale annotated dataset to facilitate the study of profile consistency identification in open-domain dialogues. We leverage the structure information in profiles to enrich the BERT representations and obtain significant performance improvements over strong baselines. We further test the proposed method on two downstream tasks. Evaluation results show the effectiveness of the proposed approach. 

We believe KvPI will be a useful resource for the research of open-domain dialogue consistency. Although there has been a lot of dialogue generation models in this field, most of them still can't understand the consistency relationship in the generation process. One of the major bottlenecks is the lack of data. Because the KvPI dataset has paired key-value profiles and dialogues, it can also be a high-quality resource for personalized dialogue generation tasks. Furthermore, because we have fine-grained consistency labels, this dataset also provides an opportunity to leverage natural language understanding models to assist dialogue generation models.
We hope that the data will aid training dialogue agents to be more consistent.

\section*{Acknowledgments}

This paper is supported by the National Natural Science Foundation of China under Grant No.62076081, No.61772153, and No.61936010. We thank all the anonymous reviewers for their helpful comments and suggestions.

\bibliographystyle{emnlp2020}
\bibliography{emnlp2020}

\end{document}